%% file: main.tex
\documentclass[sigconf]{acmart}
\setcopyright{none}
\renewcommand\footnotetextcopyrightpermission[1]{}

\usepackage{graphicx}
\usepackage{amsfonts}
\usepackage{amsmath}
\usepackage{booktabs}
\usepackage{xcolor}
\definecolor{my_green}{RGB}{114,219,190}
\definecolor{my_purple}{RGB}{181,41,183}
\definecolor{my_orange}{RGB}{162,66,17}

\usepackage{multirow}
\usepackage[normalem]{ulem}
\usepackage[super]{nth}
\usepackage{cleveref}
\usepackage{colortbl}
\usepackage{xspace}
\usepackage{caption}
\usepackage{multirow}
\usepackage{makecell}

\AtBeginDocument{%
  }

\begin{document}

\title{UniMoGen: Universal Motion Generation}
\newcommand{\name}{UniMoGen\xspace}

\author{Aliasghar Khani}
\affiliation{%
  \institution{Autodesk Research}
  \country{Canada}
  }
\email{aliasghar.khani@autodesk.com}

\author{Arianna Rampini}
\affiliation{%
  \institution{Autodesk Research}
  \country{Canada}
  }
\email{arianna.rampini@autodesk.com}

\author{Evan Atherton}
\affiliation{%
  \institution{Autodesk Research}
  \country{Canada}
  }

\author{Bruno Roy}
\affiliation{%
  \institution{Autodesk Research}
  \country{Canada}
  }

\renewcommand{\shortauthors}{Khani et al.}

\input{0-abs}

\ccsdesc[500]{Computing methodologies~Motion processing}

\keywords{Motion Generation, Diffusion models, Unet, skeleton-agnostic}
\begin{teaserfigure}
  \centering
  \includegraphics[height=5cm]{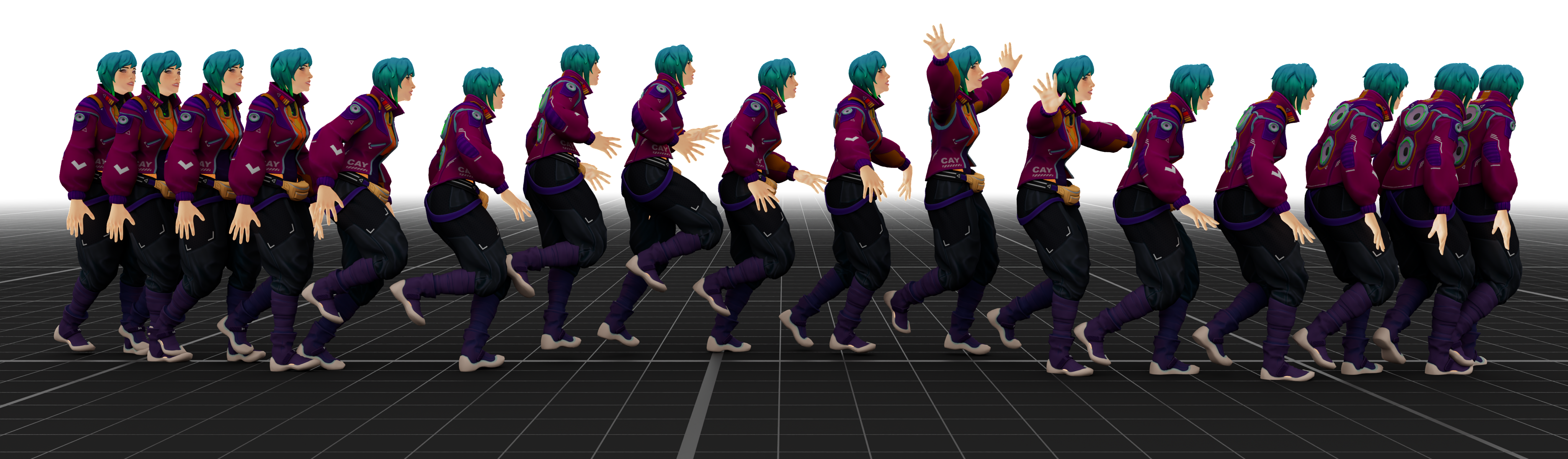}
  %\caption{A sample motion sequence generated by \name.}
  \caption{\name generates realistic and diverse character motions in real time, controllable via action type, trajectory, and past motion context. It supports arbitrary skeleton topologies by operating in a skeleton-agnostic manner, and can produce long, smooth motion sequences that transition seamlessly across different styles. The figure shows a sample motion sequence generated by \name. }
  \label{fig:teaser}
\end{teaserfigure}


\maketitle
\pagestyle{plain}

\input{1-intro}
\input{2-related}
\input{3-method}
\input{4-exp}
\input{6-conclusion}

% \begin{acks}
% \end{acks}

\bibliographystyle{ACM-Reference-Format}
\bibliography{bibliography.bib}

\input{7-figures-only}


\end{document}

%% file: 0-abs.tex
\begin{abstract}
Motion generation is a cornerstone of computer graphics, animation, gaming, and robotics, enabling the creation of realistic and varied character movements. A significant limitation of existing methods is their reliance on specific skeletal structures, which restricts their versatility across different characters. To overcome this, we introduce \name, a novel UNet-based diffusion model designed for skeleton-agnostic motion generation. \name can be trained on motion data from diverse characters, such as humans and animals, without the need for a predefined maximum number of joints. By dynamically processing only the necessary joints for each character, our model achieves both skeleton agnosticism and computational efficiency. Key features of \name include controllability via style and trajectory inputs, and the ability to continue motions from past frames. We demonstrate \name’s effectiveness on the \textsc{100style} dataset, where it outperforms state-of-the-art methods in diverse character motion generation. Furthermore, when trained on both the \textsc{100style} and \textsc{LAFAN1} datasets, which use different skeletons, \name achieves high performance and improved efficiency across both skeletons. These results highlight \name's potential to advance motion generation by providing a flexible, efficient, and controllable solution for a wide range of character animations.
\end{abstract}

%% file: 1-intro.tex
\section{Introduction}
The generation of realistic and diverse character motions is essential in various domains, including computer graphics, animation, gaming, and robotics. Motion generation enables the creation of lifelike animations that enhance user experiences in films, video games, virtual reality, and robotic simulations~\cite{holden2016deep}. Previous research has demonstrated significant progress in data-driven approaches to motion generation~\cite{motionlab, motionllama, lamp, dart, closd, camdm, mdm, survey}. However, these techniques are often tailored to specific skeletal structures, limiting their applicability to characters with different topologies. This presents a major challenge in developing a universal model capable of generating motion for a wide range of characters, such as humans, animals, and fantastical creatures, each with distinct skeletal configurations.

Recent advancements in motion generation have aimed to address the challenge of producing animations using diffusion and auto-regressive models, but limitations remain.
For instance, MDM~\cite{mdm} introduced the first motion diffusion model conditioned on text input. While pioneering, it does not incorporate trajectory information for controllability or utilize past motion frames for auto-regressive generation. 
Building on this, CAMDM~\cite{camdm} employs an auto-regressive diffusion framework with a transformer-based architecture to generate high-quality motions based on user control signals and prior motion, achieving real-time performance. Although this method improves controllability by leveraging trajectory and past motion, it passes all inputs and conditions into the transformer at once. This results in unnecessarily long input sequences, leading to increased memory usage and slower generation times. 
MotionLLaMA~\cite{motionllama}, on the other hand, proposes a transformer-based auto-regressive model that tokenizes motion sequences and, given text or audio, generates motion sequences through next-token prediction. While this approach is compatible with language modeling frameworks, applying tokenization to continuous motion data, which is highly sensitive to small value changes, can lead to subtle but important information loss~\cite{descretequant}. This degradation in precision negatively affects the quality of the generated motions. 
% consider moving some of this to the related work section
In addition to these limitations, like many other methods, these methods are designed for a single skeletal structure and require separate training for each distinct skeleton, restricting their generalizability.
In contrast, AnyTop~\cite{anytop} introduces a diffusion model capable of generating motions for arbitrary skeletons by integrating topology information into a transformer-based denoising network. However, it requires specifying a maximum number of joints in the skeletons; if a skeleton has fewer joints, the model pads them, resulting in unnecessary computational and time overhead.

To overcome these challenges, we present \name, a novel approach to motion generation that is inherently skeleton-agnostic. \name is built upon a UNet-based diffusion model with attention modules. The UNet architecture enhances efficiency by first downsampling the motion sequence in the temporal dimension and applying attention modules to the shorter sequences. Additionally, the attention modules enable \name to handle motion data from characters with varying numbers of joints without requiring padding or fixed skeletal templates. By temporally downsampling the motion sequence and processing only the relevant joints for each character, \name achieves both skeleton agnosticism and computational efficiency, making it suitable for large-scale applications.

This work introduces several key contributions that advance the field of motion generation:
\begin{itemize}
    \item \textbf{Skeleton-Agnostic Architecture:} \name is the first model to seamlessly handle arbitrary skeletal structures without padding or fixed joint counts, enabling simultaneous training on diverse characters, such as humans and animals, and setting a new standard for universal motion generation.

    \item \textbf{Efficient and Controllable Motion Synthesis:} By leveraging a UNet-based diffusion model with temporal downsampling and attention mechanisms, \name achieves high computational efficiency while offering fine-grained control through style and trajectory inputs, as well as the ability to continue motion sequences from past frames.

    \item \textbf{Real-Time Generation:} \name supports real-time motion synthesis, generating motions in just 0.09 seconds on a GPU.
\end{itemize}
% \textbf{Skeleton-Agnostic Architecture:} \name is the first model to seamlessly handle arbitrary skeletal structures without padding or fixed joint counts, enabling simultaneous training on diverse characters, such as humans and animals, and setting a new standard for universal motion generation.

% \textbf{Efficient and Controllable Motion Synthesis:} By leveraging a UNet-based diffusion model with temporal downsampling and attention mechanisms, \name achieves high computational efficiency while offering fine-grained control through style and trajectory inputs, as well as the ability to continue motion sequences from past frames.

% \textbf{Real-Time Generation:} \name supports real-time motion synthesis, generating motions in just 0.09 seconds on a GPU.

% \textbf{Empirical Superiority and Flexibility:} Extensive experiments on the 100style~\cite{100style} and HumanAct12~\cite{humanact12} datasets demonstrate that \name outperforms state-of-the-art methods like CAMDM\cite{camdm} and AnyTop~\cite{anytop} in motion quality, diversity, and efficiency, while its ability to synthesize hand motions for characters lacking them showcases its adaptability to partial skeletal data.

We evaluate \name on the \textsc{100style}~\cite{100style} dataset, a comprehensive collection of stylized human locomotion data encompassing 100 different styles, such as walking, running, and sidestepping. In this benchmark, our method outperforms MDM and CAMDM, demonstrating its superior ability to generate diverse and high-quality motions. For example, compared to CAMDM, \name reduces the percentage of frames with foot penetration from 4.73\% to 0.3\% on average for both left and right feet, and decreases the average foot sliding distance from 0.98 to 0.56.

To further test its scalability, we train \name on a combination of the \textsc{100style} and \textsc{LAFAN1}~\cite{lafan} datasets, which provide a broad spectrum of human actions, including daily activities like walking, running, and sitting down. In this more comprehensive setting, \name not only achieves better performance than AnyTop but also does so with improved efficiency, highlighting its robustness across different datasets and its potential for real-world applications.
For instance, our method achieves an average foot penetration percentage of 11.05\% (across both feet), significantly lower than AnyTop's 26.41\%.

The remainder of this paper is structured as follows: Section~\ref{related_work} provides an overview of related work in motion generation. Section~\ref{method} delves into the architectural details of \name, explaining how it achieves skeleton agnosticism and efficiency. Section~\ref{experiments} describes the datasets used, the experimental methodology, and results comparing \name with baseline methods. Finally, Section~\ref{conclusion} concludes the paper and outlines potential avenues for future research.

%% file: 2-related.tex
\section{Related work}
\label{related_work}
\subsection{Motion Generation with Diffusion Models}
Diffusion models have emerged as a powerful framework for motion generation, leveraging their ability to produce high-quality, diverse samples. For instance, MDM~\cite{mdm} was a pioneering work that adapted diffusion models for motion synthesis, generating sequences from text or style inputs without auto-regression. However, it lacks the temporal continuity typically provided by auto-regressive methods. 
Building on this direction, FlowMDM~\cite{flowmdm} introduces a transformer-based bidirectional diffusion model that generates long, smooth, and realistic human motion sequences conditioned on multiple textual descriptions. By combining a bidirectional Transformer, blended positional encodings, and pose-centric cross-attention, it effectively captures both past and future motion dependencies, enabling seamless transitions and eliminating the need for post-processing. 
Complementary to these text-driven approaches, CAMDM~\cite{camdm} focuses on real-time motion generation using motion diffusion probabilistic models. It enables high-quality and diverse character animations in response to dynamic user-supplied control signals. A notable contribution of CAMDM is its support for real-time interactive control.
Further enhancing controllability and realism, DART~\cite{dart} introduces a diffusion-based auto-regressive model that generates long human motion sequences in real time, conditioned on both text and motion history. Operating in a learned latent motion primitive space, DART supports continuous text-driven generation and allows fine-grained spatial control—such as reaching target poses or navigating to specific locations via latent noise optimization and reinforcement learning.

In contrast to our method, all these methods are designed to work with a single skeleton each time and cannot be used to train on a dataset with different skeletons. In addition to that, our method uses 1D convolutions to downsample and upsample the data in the temporal dimension, which helps in reducing attention costs. However, as these methods use transformer architecture, they keep the number of frames untouched and all attention operations are conducted at the original frame length.

\subsection{Auto-Regressive Motion Generation}
Numerous works have explored auto-regressive models that follow the next-token prediction paradigm for motion generation. For example, T2M-GPT~\cite{t2m_gpt} employs a Vector Quantized-Variational AutoEncoder (VQ-VAE) to discretize motion sequences into code indices, and a GPT-like model to perform auto-regressive next-index prediction conditioned on previous indices and a text description. Similarly, LaMP~\cite{lamp} also uses a VQ-VAE to encode motion but adopts a masked prediction strategy instead of standard auto-regression. During inference, LaMP performs iterative masked prediction: it begins with a completely masked sequence, estimates distributions for the masked tokens, samples tokens, and re-masks low-confidence tokens over multiple steps. The generation process is conditioned by motion-informative text features extracted using LaMP’s pre-trained text transformer, replacing the commonly used CLIP embeddings. Extending this direction, MotionLLaMA~\cite{motionllama} leverages a Large Language Model (LLM) fine-tuned with LoRA to handle various motion-related tasks. 
It introduces the Holistic Motion (HoMi) tokenizer to convert continuous motion into discrete tokens and performs motion generation in a unified auto-regressive framework using the causal language model (LLaMA3.2-Instruct), predicting the next motion token based on past tokens and conditioning signals such as text or audio.
% This auto-regressive, tokenizer-based approach, combined with the powerful LLM backbone and the extensive MotionHub dataset, allows MotionLLaMA to address a wide range of synthesis tasks within a single framework, including text-to-motion, music-to-dance, and motion completion, achieving state-of-the-art or comparable performance.

Like motion diffusion models and unlike our method, all these methods are designed to work with only one skeleton. Moreover, training a tokenizer for the motion data, which is a continuous one and very sensitive to small variations in values, is very challenging and will degrade the quality~\cite{descretequant}.

\subsection{Skeleton-Agnostic Motion Generation}
Addressing the long-standing challenge of generating motion for arbitrary skeletons, Gat et al. introduce AnyTop~\cite{anytop}, a diffusion model designed to generate motions for diverse characters with distinct motion dynamics using only their skeletal structure as input. This work specifically tackles the problem of handling a wide variety of skeletal topologies, including skeletons which vary significantly in structure. AnyTop utilizes a transformer-based denoising network tailored for arbitrary skeleton learning, incorporating topology information and textual joint descriptions to learn semantic correspondences across diverse skeletons. A key design choice is embedding each joint independently at each frame, enabling greater flexibility compared to methods that embed the entire pose. The model demonstrates generalization to unseen skeletons and can produce natural motions for a range of character types like bipeds, quadrupeds, and multi-legged creatures. AnyTop stands out as a skeletal-based approach capable of generating smooth motions on a diverse range of characters using a single unified model without topology-specific adjustments.

However, a key limitation of AnyTop is its reliance on a predefined maximum number of joints. For skeletons with fewer joints, it pads the joint dimension with zeros, leading to unnecessary computational and memory overhead. Conversely, if a skeleton exceeds this joint limit, the model must discard the extra joints, resulting in a loss of valuable information.

In contrast, our method leverages a U-Net architecture with attention modules that process joints independently, eliminating padding and enabling efficient training on datasets with different skeletons, such as \textsc{100style} and \textsc{LAFAN1}. Our skeleton-agnostic design, combined with auto-regressive diffusion and trajectory conditioning, allows for flexible and high-quality motion generation across varied skeletal structures, addressing the computational and data loss issues inherent in methods like AnyTop.

%% file: 3-method.tex
\section{Method}
\label{method}
In \name, our goal is to train a skeleton-agnostic motion model that combines auto-regressive generation with diffusion-based training. This design enables the model to produce arbitrarily long motion sequences while maintaining high motion quality. For a high-level overview of our diffusion architecture, please refer to Figure~\ref{fig:unimogen}.
\input{images/unimogen}

\subsection{Diffusion Models}
Diffusion models are a class of generative models that learn to reverse a gradual noising process to generate data samples~\cite{ddpm}. They operate by modeling a Markov chain that incrementally adds noise to data over a series of time steps, defined by a forward process \( q(\mathbf{x}_t | \mathbf{x}_{t-1}) \). This process transforms the original data distribution \( \mathbf{x}_0 \sim p_{\text{data}} \) into a noise distribution, typically Gaussian, at the final time step \( T \). The reverse process, parameterized by \( p_\theta(\mathbf{x}_{t-1} | \mathbf{x}_t) \), is learned to denoise the data step-by-step, starting from pure noise to reconstruct samples resembling the training data. One of the possible training objectives minimizes the difference between the clean and predicted data, often using a simplified mean-squared error loss:
\[
\mathcal{L} = \mathbb{E}_{\mathbf{x}_0, \epsilon, t} \left[ \|x_0 - \hat{x}_0(\mathbf{x}_t, t)\|^2 \right],
\]
where \( x \) is the true noise, and \( \hat{x}_0 \) is the model's prediction. This framework has shown remarkable success in generating high-quality samples across various domains, including images~\cite{ddpm, stable_diffusion} and time-series data~\cite{ts_diffusion_survey, camdm, mdm}, due to its stable training dynamics and ability to capture complex data distributions.

\subsection{Universal Motion Generation}
We propose a novel auto-regressive diffusion model for motion generation, designed to be agnostic to skeleton structures, enabling simultaneous training across diverse skeleton types. 
% Our model, referred to as \name, takes as input a style index from a predefined set of styles, a diffusion time step, a trajectory comprising \( F \) motion frames (with position trajectory and root rotation trajectory), and optionally, \( F' \) past frames (including root positions and joint rotations). Then, it outputs joint rotations and root positions for \( F \) motion frames.
Our model, referred to as \name, takes as input a style index, a diffusion time step, a trajectory, and optionally, past frames. The style index is selected from a predefined set of styles. The trajectory consists of \( F \) motion frames, including position and root rotation trajectories. The past frames, when provided, contain \( F' \) motion frames comprising root positions and joint rotations. The model then outputs joint rotations and root positions for \( F \) motion frames.

The architecture of \name is based on a U-Net with 1D convolutions along the temporal dimension. In addition, it incorporates attention modules across both temporal and joint dimensions, cross-attention modules to inject trajectory information, and Feature-wise Linear Modulation (FiLM)~\cite{film} to condition on time and style. FiLM works by passing the time step and style index through linear layers to get scale and shift parameters. These are then used to modulate the normalized features following Group Normalization~\cite{group_norm}, allowing the model to dynamically adapt its behavior based on the temporal and stylistic context.

In the joint attention module, we compute an attention mask based on the skeleton topology, restricting each joint to attend only to its ancestors, thus preserving kinematic constraints. Following the U-Net paradigm, the encoder downsamples the temporal dimension, and the decoder upsamples it, allowing the temporal attention module to operate on shorter sequences, which reduces computational overhead.

During training, the model receives the following inputs: style index \( S \in \mathbb{R} \), time step \( t \in \mathbb{R} \), position trajectory \( T_p \in \mathbb{R}^{F\times 2} \), root rotation trajectory \( T_r \in \mathbb{R}^{F\times6} \), past root positions \( P_p \in \mathbb{R}^{F'\times 3} \), past joint rotations \( P_r \in \mathbb{R}^{F'\times J\times 6} \), current root positions \( C_p \in \mathbb{R}^{F\times 3} \), and current joint rotations \( C_r \in \mathbb{R}^{F\times J\times 6} \), where \(J\) is the number of joints of a skeleton.
To train the model, Gaussian noise is added to \( C_p \) and \( C_r \) according to the diffusion schedule.
These noisy versions are then concatenated with \(P_p\) and \(P_r\), respectively, and the model is trained to denoise them. Conditioning is applied in two ways: the model uses cross-attention to incorporate information from the trajectory inputs \(T_p\) and \(T_r\), while FiLM layers condition the model on the style index \(S\) and time step \(S\).

As seen in the representations of \( P_r \) and \( C_r \), the joints are kept as a separate dimension and are processed by the joint-wise attention mechanism that supports variable-length inputs (i.e., varying numbers of joints) and facilitates information sharing across joints. This design enables the model to accommodate skeletons with different joint counts, eliminating the need for padding and avoiding unnecessary computational overhead, thereby ensuring efficient handling of diverse skeletal structures.

The loss functions used during training include the diffusion loss \(\mathcal{L}_{\text{d}}\); mean squared error (MSE) between the predicted and ground truth angular velocity of joint rotations, denoted as \(\mathcal{L}_{\text{av}}\); MSE between the ground truth and predicted global position of joints, \(\mathcal{L}_{\text{gp}}\); and MSE between the ground truth and predicted velocity of global position of joints, \(\mathcal{L}_{\text{vgp}}\). In addition, we include a foot contact loss \(\mathcal{L}_{\text{foot}}\) , defined as the \(L_2\) norm of the predicted global velocity of toe joints on frames where those joints are in contact with the ground. In the ablation study section, we demonstrate the effectiveness of combining these auxiliary losses with the original diffusion loss.

Finally, to support auto-regressive generation, \name reuses the last \(F'\) frames of the generated motion as the past frames for the next generation step. This mechanism enables the model to produce temporally coherent motion sequences of arbitrary length by chaining together successive predictions.

\subsection{Implementation Details}
Our U-Net architecture consists of three layers in both the encoder and decoder. Each layer doubles the number of feature channels and reduces the temporal resolution by half, except for the final encoder layer and the first decoder layer, which preserve the temporal resolution.

During training, we drop \(S\) with a probability of \(10\%\) to enable Classifier-Free Guidance~\cite{cfg} during inference. Additionally, we drop \(P_p\) and \(P_r\) with a probability of \(50\%\) to allow the model to learn both to generate motion from scratch, without any past context, and to continue an existing motion sequence when past frames are provided. Furthermore, we apply a Gaussian filter to the trajectory positions at random and occasionally rotate the entire motion path to encourage robustness and invariance to trajectory transformations.

For the diffusion process, we adopt a cosine beta scheduler with 50 steps for the DDPM training phase and 4 steps for DDIM during inference, balancing quality and efficiency. Optimization is performed using the Adam optimizer with a learning rate of \(1\times10^{-4}\) , along with an exponential learning rate decay where the decay factor (gamma) is set to \(0.9999\). We trained both experiments on \(8\times \text{H}100\) GPUs: 
\(34\)K steps for the \textsc{100Style} dataset and \(164\)K steps for the combined \textsc{100Style} and \textsc{LAFAN1} datasets.

%% file: images/unimogen.tex
\begin{figure*}[!t]
    \centering
    \includegraphics[width=\textwidth,trim=30pt 0 0 0,clip]{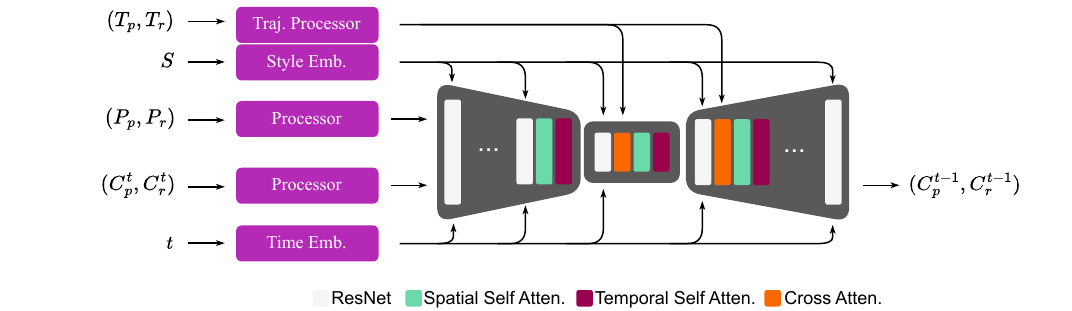}
    \caption{\textbf{Overview of the \name denoising architecture.} 
    During training, the model receives style index $S$, past motion inputs as root positions $P_p$ and joint rotations $P_r$, trajectory $(T_p, T_r)$, and diffusion time step $t$. 
    Dedicated modules process each input, and their representations are fused in a UNet-based diffusion network. The network leverages temporal and joint-level self-attention, cross-attention to inject trajectory information, and Feature-wise Linear Modulation (FiLM) \cite{film} to condition on time and style. 
    The model outputs future motion $(C_p, C_r)$, enabling controllable, skeleton-agnostic generation across diverse characters. As illustrated in the figure, we omit attention modules in the first and last layers of the UNet and apply them only to the downsampled layers to reduce memory consumption.}
    \label{fig:unimogen}
    \Description{__}
\end{figure*}

%% file: 4-exp.tex
\section{Experiments}
\label{experiments}
In this section, we evaluate the performance of \name through a series of experiments designed to assess its motion generation quality, physical plausibility, and skeleton-agnostic capabilities. We describe the datasets used, the evaluation metrics, the baseline methods for comparison, quantitative and qualitative results, and an ablation study to analyze key design choices.

\subsection{Dataset}
We evaluate our method using two diverse motion capture datasets, \textsc{100style}~\cite{100style} and \textsc{LAFAN1}~\cite{100style}, each characterized by distinct skeleton structures (i.e., a single skeleton type per dataset). The \textsc{100style} dataset comprises \(1,372\) clips, totaling \(4,094,607\) frames, and encompasses \(100\) distinct styles. On the other hand, the \textsc{LAFAN1} dataset includes \(1540\) clips with a total of \(978,844\) frames and \(15\) styles. For both training and evaluation, we utilize these datasets in their original forms without retargeting, preserving their original skeleton configurations. For motion representation, we adopt the 6D rotation representation proposed by~\cite{6drepresentation} for joint rotations, along with 3D root positions. Prior to training, we apply min-max normalization to the root position data, scaling it to the range \([-1, 1]\). The rotation data, however, is left unnormalized since it already falls within the same range.
For both datasets, we split the data into training, validation, and test sets with \(75\%\), \(15\%\), and \(10\%\) of the clips, respectively, ensuring that the style distribution is preserved across all splits.
The validation set is used for ablation studies and selecting the best model checkpoint, while the test set is reserved for final comparisons with other methods.

\subsection{Metrics}
To demonstrate the effectiveness of \name, we employ a comprehensive set of evaluation metrics. First, we use the Fréchet Inception Distance (\emph{FID}) to measure the distributional similarity between ground truth and generated motion sequences. Following prior work \cite{camdm}, we train a motion classifier on joint positions from the training set and use its feature activations to embed both real and generated motions. FID is then computed as the Fréchet distance between the resulting feature distributions.
In addition to FID, we compute two diversity metrics designed to quantify variation in generated motion. The \emph{Diversity (intra-motion)} measures the variance of each joint's spatial location over time within a single sequence, averaged across all joints and motions. The \emph{Diversity (inter-motion)} measures the variance of joint positions across different motions, averaged across all joints.

To further evaluate realism, we compute \emph{Foot Penetration} and \emph{Foot Sliding}.
%Second, we assess foot penetration, which 
Foot penetration quantifies the fraction of frames in which any of the toes (left or right) intersects the ground, indicating physical implausibility. 
%Third, we compute the foot sliding distance, a critical metric for evaluating motion realism. 
The foot sliding distance is a critical metric for evaluating motion realism, which measures the moving distance (in meters) of the character’s toes when the joint height is below a threshold (0.01 m), capturing unnatural sliding artifacts. 
Finally, we evaluate \emph{Trajectory Distance}, which comprises two components: the position difference, measured as the distance between the trajectory and the root joint positions along the $x$ and $z$ axes; and the rotation difference, which compares the root joint’s orientation with the trajectory's target rotation.

\subsection{Baselines}
We compare \name against several state-of-the-art approaches. First, we consider MDM~\cite{mdm}, the pioneering diffusion-based motion generation model that generates motion solely from text or style inputs and is not auto-regressive. Second, we include CAMDM~\cite{camdm}, a transformer-based auto-regressive diffusion model that generates motion sequences conditioned on style, trajectory, and past motion, representing the current state-of-the-art in motion generation using trajectory, style, and past motion. Lastly, as \name is skeleton-agnostic, we compare it with AnyTop~\cite{anytop}, a skeleton-agnostic motion generation method designed to handle diverse skeleton structures.

All baseline methods were trained until convergence. On the \textsc{100Style} dataset, this required 400k steps for CAMDM, 324k for MDM, and 34k for our method. When training on the combination of \textsc{100Style}~\cite{100style} and \textsc{LAFAN1}~\cite{100style} datasets, convergence was reached after 176K steps for AnyTop and 164K steps for our method.

% To compare our method with CAMDM and MDM, we trained CAMDM for 400K steps, MDM for 324K steps, and \name for only 34K steps, demonstrating that our model requires significantly less training. 
% Similarly, for the comparison with AnyTop, we trained AnyTop for 176K steps and our method for 164K steps.

\subsection{Results}
To highlight the performance of \name, we present both quantitative and qualitative results. 
For a fair comparison with MDM, which only conditions on style, we generate $500$ samples per style. Similarly, using our method, we generate $500$ samples per style by randomly selecting (past motion, trajectory) pairs corresponding to that style from the test set. In contrast, for CAMDM, we use the entire test set and generate one sample for each (style, past motion, trajectory) pair using both \name and CAMDM.
%\name consistently outperforms both baselines, achieving lower FID scores, fewer foot penetration frames, and reduced sliding distances, indicating superior motion quality and physical plausibility.Table~\ref{tab:trajectory_errors} further compares \name with CAMDM on trajectory errors (position and rotation) on the \textsc{100style} dataset, demonstrating that \name more accurately adheres to the provided trajectories. 
Table~\ref{tab:100style_metrics} compares \name with MDM and CAMDM on the \textsc{100style} dataset across generation metrics (FID and diversity) and physical plausibility metrics (number of foot penetration frames and foot sliding distance).

% \textcolor{red}{after adding root position variance this part should be revised} 

The two top rows show that, despite operating with only 4 denoising steps (i.e., 250 times fewer steps than those required by MDM), \name achieves a significantly lower FID than MDM, reflecting a meaningful enhancement in distribution alignment and perceptual quality. 
Additionally, \name is conditioned on past frames and trajectory, which increases the complexity of the task: satisfying multiple, potentially conflicting or difficult-to-model constraints simultaneously, such as style, precise past frames, a specific future trajectory, and maintaining physical plausibility, is more challenging than generating plausible motion with fewer constraints, as MDM does. Finally, the substantial improvement in diversity highlights that, unlike MDM which often produces static or repetitive motion, \name is able to generate a broader and more expressive set of styles while remaining faithful to the input signals.
Meanwhile, \name outperforms CAMDM in diversity as well as all physical plausibility metrics, achieving lower foot penetration and reduced sliding, shown in the bottom two rows. For visual examples, refer to Figure~\ref{fig:foot_contact}. 

%Although MDM shows slightly better contact metrics and diversity, but it uses $1000$ inference steps, $250 \times$ more than \name, resulting in impractical latency for interactive applications. In contrast, \name generates each sample in $4$ seconds on CPU, and in 0.09 seconds on GPU. 
Given that CAMDM is positioned as MDM’s real-time counterpart, \name demonstrates a superior trade-off between quality and efficiency, enabling physically plausible motion synthesis in real-time, while additionally supporting skeleton-agnostic generation. Each motion generation by \name takes $4$ seconds on a CPU and only $0.09$ seconds on a GPU.
\input{tables/camdm_mdm}
\input{tables/camdm_traj}

A key advantage of \name is its ability to train on multiple skeleton types simultaneously, enabling the development of a large, universal model capable of handling diverse skeletal structures without modification. To evaluate this capability, we conduct a cross-dataset comparison between \name and AnyTop using a combined dataset consisting of \textsc{100style} and \textsc{LAFAN1}, two datasets with distinct skeleton types. Table~\ref{tab:combined_metrics} reports quantitative results for this comparison. As shown, \name outperforms AnyTop across all metrics, despite not using a text encoder to encode joint names. Furthermore, our method is more efficient than AnyTop as it avoids joint padding, which introduces unnecessary computational and time overhead.  Instead, \name maintains joints as a separate dimension and leverages attention mechanisms to efficiently process varying skeletal structures. As shown in Figure~\ref{fig:multi_skeleton}, this approach enables \name to generate motions across different skeleton types. The left and right panels illustrate motions generated for the skeletons of the \textsc{LAFAN1} and \textsc{100Style} datasets, respectively.

For more qualitative results and comparisons, please refer to the supplemental video.

\input{tables/anytop}

\paragraph{Style Blending.}
To further illustrate the flexibility of \mbox{\name's} style conditioning, we present a style blending experiment, where we interpolate between two style embeddings (e.g., \(30\%\) Style A and \(70\%\) Style B). As shown in Figure~\ref{fig:style_blending}, the generated motions smoothly transition between the characteristics of both styles, demonstrating the continuous and expressive nature of the style embedding space learned by our model.
Style blending animations examples are also included in the supplemental video.

% \textcolor{red}{We refer to the supplemental video for examples of style blending and additional results.}

\input{images/style_blending}

\subsection{Ablation Study}
% We conduct an ablation study to investigate the impact of key components and design choices in \name. Specifically, we examine the following: (1) the effect of min-max normalization on root positions, assessing its necessity for stable training; (2) the use of a cosine scheduler with fewer diffusion steps, evaluating its impact on generation quality and computational efficiency; (3) the use of separate attention modules for spatial and temporal dimensions versus a single attention module that merges both dimensions before processing, showing the power of separate attentions; (4) the inclusion of positional encoding (as used in transformer models) to capture temporal relationships; (5) the effect of dataset balancing to ensure equitable representation of styles and skeletons; (6) the role of auxiliary losses in enhancing output quality. These ablations provide insights into the critical components driving the performance of \name.
In this part, we conduct ablation studies to evaluate the impact of key design choices in \name, reporting FID, Penetration (Frames), and Sliding (m) on the validation set of the \textsc{100style} dataset. First, in Table~\ref{tab:ablation_norm}, we assess the effect of min-max normalization on root positions, a preprocessing step to stabilize training by scaling data to a fixed range. An analysis of the raw data reveals that while all six components of the joint rotation representation fall within the range \([-1,1]\), the X, Y, and Z components of the root position vary significantly in scale. Specifically, the ranges for X, Y, and Z are \([-3.52, 3.63]\), \([0.77, 1.21]\), and \([-2.91, 4.01]\), respectively, with the Y-axis exhibiting the smallest variation. This imbalance hinders effective learning of the Y-axis component, contributing to increased foot penetration errors. To mitigate this issue, we apply min-max normalization to the X, Y, and Z components of the root position and compare model performance with and without normalization to evaluate its role in ensuring stable convergence.
\input{tables/abl-min-max-norm}

Second, we evaluate the use of a cosine noise scheduler with fewer diffusion steps, to balance generation quality and computational efficiency, testing 50 steps against the standard 1000 steps. As shown in  Table~\ref{tab:ablation_scheduler}, we can achieve both better results and faster generations using cosine scheduler. 
\input{tables/abl-cosine-sched}

Third, we compare separate attention modules for spatial (joint) and temporal dimensions, which allow specialized feature processing, against a single attention module that merges both dimensions before processing. The results in Table~\ref{tab:ablation_attention} demonstrate the advantage of decoupled attention mechanisms.
\input{tables/abl-sep-attn}

Fourth, we investigate the inclusion of positional encoding, as used in transformer models~\cite{attention}, to capture positional relationships in both temporal and spatial (joints) dimensions, testing its impact on motion coherence.  Table~\ref{tab:ablation_posenc} shows that including positional encodings leads to improved performance.
\input{tables/abl-pos-enc}

Fifth, we examine dataset balancing to ensure equitable representation of styles, mitigating bias toward overrepresented categories. This strategy yields improvements across metrics, as reported in Table~\ref{tab:ablation_balance}.
\input{tables/abl-data-balance}

Finally, we analyze the role of auxiliary losses, which regularize training and enhance output quality. Table~\ref{tab:ablation_auxloss} confirms that including auxiliary losses improves the overall results.
\input{tables/abl-aux-loss}

%% file: tables/camdm_mdm.tex
% \begin{table}[h]
% \centering
% \caption{Comparison on 100style dataset: FID, foot penetration frames, and foot sliding distance.}
% \label{tab:100style_metrics}
% \begin{tabular}{lccc}
% \toprule
% Method & FID $\downarrow$ & Foot Penetration (Frames) $\downarrow$ & Foot Sliding Distance (m) $\downarrow$ \\
% \midrule
% MDM & 5.23 & 1243 & 0.87 \\
% CAMDM & 3.91 & 892 & 0.65 \\
% \name & \textbf{2.84} & \textbf{576} & \textbf{0.42} \\
% \bottomrule
% \end{tabular}
% \end{table}

\begin{table}[h]
\centering
\caption{\textbf{Comparison with MDM and CAMDM on the \textsc{100style} dataset.}
Compared to MDM, our model operates with 250x fewer inference steps and is more controllable by conditioning on past frames and trajectory. Despite these constraints, it outperforms MDM in terms of FID and diversity and achieves comparable overall results for foot penetration and sliding. Compared to CAMDM, our method shows superior performance across all metrics, with CAMDM having a slight advantage only in FID.}
\label{tab:100style_metrics}
\resizebox{\columnwidth}{!}{%
\begin{tabular}{lcccccc}
\toprule
% Method & FID $\downarrow$ & Diversity $\uparrow$ & Left Penetration (Frames)\% $\downarrow$ & Right Penetration (Frames)\% $\downarrow$ & Sliding (m) $\downarrow$ \\
\multirow{2}{*}[0.6em]{Method} & 
\multirow{2}{*}[0.6em]{FID $\downarrow$} & 
\makecell{Diversity $\uparrow$ \\(intra-motion)} & 
\makecell{Diversity $\uparrow$ \\(inter-motion)} & 
\makecell{Left Pen. $\downarrow$ \\(Frames)\%} & 
\makecell{Right Pen. $\downarrow$  \\(Frames)\%} & 
\makecell{Ft. Slid. $\downarrow$ \\(m)} \\
\midrule
MDM & 2.64 &0.026 &0.083 & \textbf{0.12}  & \textbf{0.15} & \textbf{0.41} \\
\name & \textbf{2.22} &\textbf{0.078} & \textbf{0.213} &  0.30 & 0.36 & 0.61 \\
% MDM & 2.6276 & 36.85 & \textbf{0.12}  & \textbf{0.15} & \textbf{0.41} \\
% \name & 2.2408 & 33.47 &  0.30 & 0.36 & 0.61 \\
\midrule
CAMDM & \textbf{2.20} & 0.052 & 0.161 &  4.73 & 4.73 & 0.98 \\
\name & 2.24 & \textbf{0.078} & \textbf{0.213} & \textbf{0.26} & \textbf{0.35} & \textbf{0.56} \\
\bottomrule
\end{tabular}%
}
\end{table}

%% file: tables/camdm_traj.tex
% \begin{table}[h]
% \centering
% \caption{Trajectory errors on 100style dataset: position and rotation differences.}
% \label{tab:trajectory_errors}
% \resizebox{\columnwidth}{!}{%
% \begin{tabular}{lcc}
% \toprule
% Method & Mean Position Error (m) $\downarrow$ & Mean Rotation Error (deg) $\downarrow$ \\
% \midrule
% CAMDM & 0.0656 & 8.1332 \\
% \name & \textbf{0.0089} & \textbf{6.9889} \\
% \bottomrule
% \end{tabular}
% }
% \end{table}

\begin{table}[h]
\centering
\caption{\textbf{Comparison of trajectory errors with CAMDM on the \textsc{100style} dataset.}
This table reports trajectory-following accuracy. Our method consistently outperforms CAMDM, demonstrating more precise adherence to the given trajectories.}
\label{tab:trajectory_errors}
\resizebox{\columnwidth}{!}{%
\begin{tabular}{lcc}
\toprule
Method & Mean Position Error (m) $\downarrow$ & Mean Rotation Error (deg) $\downarrow$ \\
\midrule
CAMDM & 0.07 & 8.13 \\
\name & \textbf{0.01} & \textbf{6.99} \\
\bottomrule
\end{tabular}
}
\end{table}

%% file: tables/anytop.tex
\begin{table}[h]
\centering
\caption{\textbf{Comparison with AnyTop on the combined \textsc{100style} and \textsc{LAFAN1} datasets.}
The first two rows report results on \textsc{100style}, while the following rows correspond to \textsc{LAFAN1}. As the results indicate, \name consistently outperforms AnyTop across both datasets by a substantial margin.}
\label{tab:combined_metrics}
\resizebox{\columnwidth}{!}{%
% \begin{tabular}{lccccc}
% \toprule
% Method & FID $\downarrow$ & Diversity $\uparrow$ & Left Penetration (Frames)\% $\downarrow$ & Right Penetration (Frames)\% $\downarrow$ & Sliding (m) $\downarrow$ \\
% \midrule
% AnyTop & 14.69 & 0.70 & 26.66 & 19.69 & 1.49 \\
% \name & \textbf{2.191} & \textbf{31.6} & \textbf{12.64} & \textbf{10.90} & \textbf{1.21}  \\
% \midrule
% AnyTop & 4.197 & 0.33 & 19.75 & 33.06 & 1.81 \\
% \name & \textbf{1.423} &  \textbf{36.4} & \textbf{9.24} & \textbf{12.85} & \textbf{1.818} \\
% \bottomrule
% \end{tabular}
\begin{tabular}{lcccccc}
\toprule
%Method & FID $\downarrow$ & Div. (intra-motion) $\uparrow$ & Div. (inter-motion) $\uparrow$ & Left pen. \% $\downarrow$ & Right Pen. \% $\downarrow$ & Ft. Slid. (m) $\downarrow$ \\
\multirow{2}{*}[0.6em]{Method} & 
\multirow{2}{*}[0.6em]{FID $\downarrow$} & 
\makecell{Diversity $\uparrow$ \\(intra-motion)} & 
\makecell{Diversity $\uparrow$ \\(inter-motion)} & 
\makecell{Left Pen. $\downarrow$ \\(Frames)\%} & 
\makecell{Right Pen. $\downarrow$  \\(Frames)\%} & 
\makecell{Ft. Slid. $\downarrow$ \\(m)} \\
\midrule
AnyTop & 14.69 & 1.31e-5 & 3.96e-5 & 26.66 & 19.69 & 1.49 \\
\name & \textbf{2.191} & \textbf{0.08} & \textbf{0.23} & \textbf{12.64} & \textbf{10.90} & \textbf{1.21}  \\
\midrule
AnyTop & 4.197 & 7.42e-5 & 2.03e-4 & 19.75 & 33.06 & 1.81 \\
\name & \textbf{1.423} & \textbf{0.14} & \textbf{0.37} & \textbf{9.24} & \textbf{12.85} & \textbf{1.82} \\
\bottomrule
\end{tabular}
}
\end{table}

%% file: images/style_blending.tex
\begin{figure}[t]
    \centering
    \includegraphics[width=\linewidth]{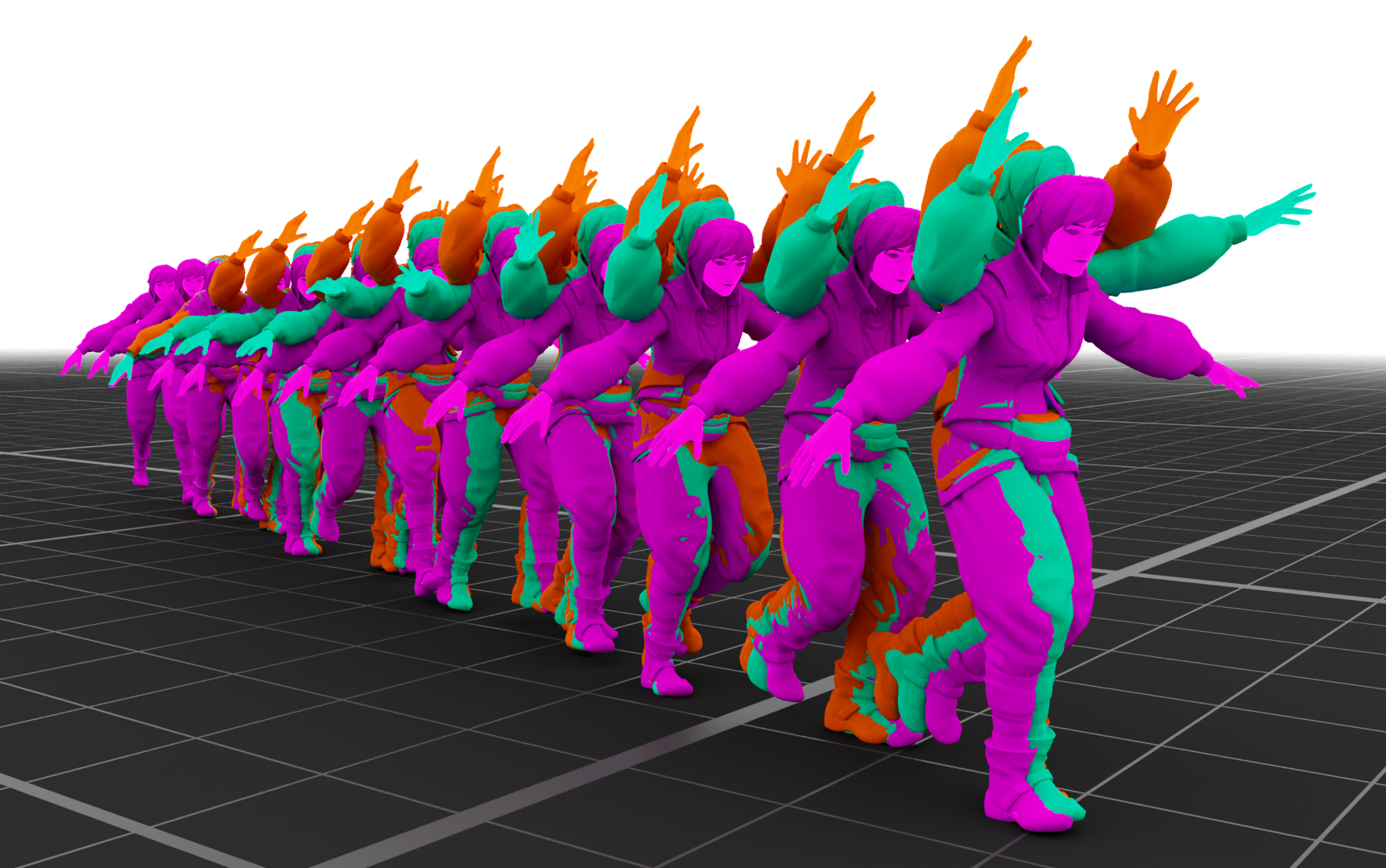}
    \caption{\textbf{Style blending with \name.} Visualization of motions generated by blending two styles: Aeroplane and Arms Above Head. \textcolor{my_purple}{Purple} shows \(100\%\) Aeroplane and \(0\%\) Arms Above Head, \textcolor{my_green}{Green} shows a blend of \(35\%\) Aeroplane and \(65\%\) Arms Above Head, and \textcolor{my_orange}{Orange} shows \(0\%\) Aeroplane and \(100\%\) Arms Above Head. The smooth transition illustrates the expressive and continuous nature of the learned style space.}
    \label{fig:style_blending}
\end{figure}

%% file: tables/abl-min-max-norm.tex
% \begin{table}[h]
% \centering
% \caption{Ablation study on min-max normalization of root positions.}
% \label{tab:ablation_norm}
% \resizebox{\columnwidth}{!}{%
% \begin{tabular}{lcccc}
% \toprule
% Configuration & FID $\downarrow$ & Left Penetration (Frames)\% $\downarrow$ & Right Penetration (Frames)\% $\downarrow$ & Sliding (m) $\downarrow$ \\
% \midrule
% W Min-Max Norm & 2.3056 & \textbf{0.0347} & \textbf{0.0375} & \textbf{0.5324} \\
% W/O Min-Max Norm & 2.2586 & 0.0757 & 0.0875 & 0.5976 \\
% \bottomrule
% \end{tabular}%
% }
% \end{table}

\begin{table}[h]
\centering
\caption{\textbf{Min-Max Normalization of Root Positions.} Normalizing root position values using min-max scaling leads to improved performance.}
\label{tab:ablation_norm}
\resizebox{\columnwidth}{!}{%
\begin{tabular}{lcccc}
\toprule
%Configuration & FID $\downarrow$ & Left Penetration (Frames)\% $\downarrow$ & Right Penetration (Frames)\% $\downarrow$ & Sliding (m) $\downarrow$ \\
Configuration & FID $\downarrow$ & Left Pen. (Frames)\% $\downarrow$ & Right Pen. (Frames)\% $\downarrow$ & Sliding (m) $\downarrow$ \\
% \multirow{2}{*}[0.6em]{Configuration} & 
% \multirow{2}{*}[0.6em]{FID $\downarrow$} & 
% \makecell{Left Pen. $\downarrow$ \\(Frames)\%} & 
% \makecell{Right Pen. $\downarrow$  \\(Frames)\%} & 
% \makecell{Ft. Slid. $\downarrow$ \\(m)} \\
\midrule
W Min-Max Norm & 2.31 & \textbf{3.47} & \textbf{3.75} & \textbf{0.53} \\
W/O Min-Max Norm & \textbf{2.26} & 7.57 & 8.75 & 0.60 \\
\bottomrule
\end{tabular}%
}
\end{table}

%% file: tables/abl-cosine-sched.tex
% \begin{table}[h]
% \centering
% \caption{Ablation study on cosine scheduler with fewer diffusion steps.}
% \label{tab:ablation_scheduler}
% \resizebox{\columnwidth}{!}{%
% \begin{tabular}{lcccc}
% \toprule
% Configuration & FID $\downarrow$ & Left Penetration (Frames)\% $\downarrow$ & Right Penetration (Frames)\% $\downarrow$ & Sliding (m) $\downarrow$ \\
% \midrule
% Cos. Sched. (100) & \textbf{2.2431} & \textbf{0.0099} & \textbf{0.0132} & \textbf{0.4936} \\
% Lin. Sched. (1000) & 2.3056 & 0.0347 & 0.0375 & 0.5324 \\
% \bottomrule
% \end{tabular}%
% }
% \end{table}

\begin{table}[h]
\centering
\caption{\textbf{Cosine Scheduler with 100 Diffusion Steps.} Employing a cosine scheduler enables effective training with only 100 diffusion steps, resulting in improved performance despite the reduced step count.}
\label{tab:ablation_scheduler}
\resizebox{\columnwidth}{!}{%
\begin{tabular}{lcccc}
\toprule
%Configuration & FID $\downarrow$ & Left Penetration (Frames)\% $\downarrow$ & Right Penetration (Frames)\% $\downarrow$ & Sliding (m) $\downarrow$ \\
Configuration & FID $\downarrow$ & Left Pen. (Frames)\% $\downarrow$ & Right Pen. (Frames)\% $\downarrow$ & Sliding (m) $\downarrow$ \\
\midrule
Cos. Sched. (100) & \textbf{2.24} & \textbf{0.99} & \textbf{1.32} & \textbf{0.49} \\
Lin. Sched. (1000) & 2.31 & 3.47 & 3.75 & 0.53 \\
\bottomrule
\end{tabular}%
}
\end{table}

%% file: tables/abl-sep-attn.tex
% \begin{table}[h]
% \centering
% \caption{Ablation study on separate vs. merged attention modules.}
% \label{tab:ablation_attention}
% \resizebox{\columnwidth}{!}{%
% \begin{tabular}{lcccc}
% \toprule
% Configuration & FID $\downarrow$ & Left Penetration (Frames)\% $\downarrow$ & Right Penetration (Frames)\% $\downarrow$ & Sliding (m) $\downarrow$ \\
% \midrule
% Separate Spatial/Temporal & \textbf{2.2963} & \textbf{0.0088} & \textbf{0.0065} & \textbf{0.4749} \\
% Merged Attention & 2.3056 & 0.0347 & 0.0375 & 0.5324 \\
% \bottomrule
% \end{tabular}%
% }
% \end{table}

\begin{table}[h]
\centering
\caption{\textbf{Separate vs. Merged Attention Modules.} Utilizing separate attention modules enhances both performance and computational efficiency compared to merged attention.}
\label{tab:ablation_attention}
\resizebox{\columnwidth}{!}{%
\begin{tabular}{lcccc}
\toprule
Configuration & FID $\downarrow$ & Left Pen. (Frames)\% $\downarrow$ & Right Pen. (Frames)\% $\downarrow$ & Sliding (m) $\downarrow$ \\
\midrule
Separate Spatial/Temporal & \textbf{2.30} & \textbf{0.88} & \textbf{0.65} & \textbf{0.47} \\
Merged Attention & 2.31 & 3.47 & 3.75 & 0.53 \\
\bottomrule
\end{tabular}%
}
\end{table}

%% file: tables/abl-pos-enc.tex
% \begin{table}[h]
% \centering
% \caption{Ablation study on positional encoding.}
% \label{tab:ablation_posenc}
% \resizebox{\columnwidth}{!}{%
% \begin{tabular}{lcccc}
% \toprule
% Configuration & FID $\downarrow$ & Left Penetration (Frames)\% $\downarrow$ & Right Penetration (Frames)\% $\downarrow$ & Sliding (m) $\downarrow$ \\
% \midrule
% With Positional Encoding & 2.2980 & \textbf{0.0114} & \textbf{0.0104} & \textbf{0.4750} \\
% Without Positional Encoding & 2.3056 & 0.0347 & 0.0375 & 0.5324 \\
% \bottomrule
% \end{tabular}%
% }
% \end{table}

\begin{table}[h]
\centering
\caption{\textbf{Positional Encoding.} As expected, incorporating positional encodings leads to improved performance in our model.}
\label{tab:ablation_posenc}
\resizebox{\columnwidth}{!}{%
\begin{tabular}{lcccc}
\toprule
%Configuration & FID $\downarrow$ & Left Penetration (Frames)\% $\downarrow$ & Right Penetration (Frames)\% $\downarrow$ & Sliding (m) $\downarrow$ \\
Configuration & FID $\downarrow$ & Left Pen. (Frames)\% $\downarrow$ & Right Pen. (Frames)\% $\downarrow$ & Sliding (m) $\downarrow$ \\
\midrule
With Positional Encoding & \textbf{2.30} & \textbf{1.14} & \textbf{1.04} & \textbf{0.48} \\
Without Positional Encoding & 2.31 & 3.47 & 3.75 & 0.53 \\
\bottomrule
\end{tabular}%
}
\end{table}

%% file: tables/abl-data-balance.tex
% \begin{table}[h]
% \centering
% \caption{Ablation study on dataset balancing.}
% \label{tab:ablation_balance}
% \resizebox{\columnwidth}{!}{%
% \begin{tabular}{lcccc}
% \toprule
% Configuration & FID $\downarrow$ & Left Penetration (Frames)\% $\downarrow$ & Right Penetration (Frames)\% $\downarrow$ & Sliding (m) $\downarrow$ \\
% \midrule
% Balanced Dataset & 2.3276 & \textbf{0.0198} & \textbf{0.0221} & \textbf{0.5159} \\
% Unbalanced Dataset & 2.3056 & 0.0347 & 0.0375 & 0.5324 \\
% \bottomrule
% \end{tabular}%
% }
% \end{table}

\begin{table}[h]
\centering
\caption{\textbf{Dataset Balancing.} Balancing the dataset across styles through oversampling yields better motions.}
\label{tab:ablation_balance}
\resizebox{\columnwidth}{!}{%
\begin{tabular}{lcccc}
\toprule
%Configuration & FID $\downarrow$ & Left Penetration (Frames)\% $\downarrow$ & Right Penetration (Frames)\% $\downarrow$ & Sliding (m) $\downarrow$ \\
Configuration & FID $\downarrow$ & Left Pen. (Frames)\% $\downarrow$ & Right Pen. (Frames)\% $\downarrow$ & Sliding (m) $\downarrow$ \\
\midrule
Balanced Dataset & 2.33 & \textbf{1.98} & \textbf{2.21} & \textbf{0.52} \\
Unbalanced Dataset & \textbf{2.31 }& 3.47 & 3.75 & 0.53 \\
\bottomrule
\end{tabular}%
}
\end{table}

%% file: tables/abl-aux-loss.tex
% \begin{table}[h]
% \centering
% \caption{Ablation study on auxiliary losses.}
% \label{tab:ablation_auxloss}
% \resizebox{\columnwidth}{!}{%
% \begin{tabular}{lcccc}
% \toprule
% Configuration & FID $\downarrow$ & Left Penetration (Frames)\% $\downarrow$ & Right Penetration (Frames)\% $\downarrow$ & Sliding (m) $\downarrow$ \\
% \midrule
% With Auxiliary Losses & 2.3056 & \textbf{0.0347} & \textbf{0.0375} & \textbf{0.5324} \\
% Without Auxiliary Losses & 2.2725 & 0.0577 & 0.0623 & 0.8085 \\
% \bottomrule
% \end{tabular}%
% }
% \end{table}

\begin{table}[h]
\centering
\caption{\textbf{Auxiliary Losses.} Incorporating auxiliary losses significantly enhances the physical plausibility of the generated motions.}
\label{tab:ablation_auxloss}
\resizebox{\columnwidth}{!}{%
\begin{tabular}{lcccc}
\toprule
%Configuration & FID $\downarrow$ & Left Penetration (Frames)\% $\downarrow$ & Right Penetration (Frames)\% $\downarrow$ & Sliding (m) $\downarrow$ \\
Configuration & FID $\downarrow$ & Left Pen. (Frames)\% $\downarrow$ & Right Pen. (Frames)\% $\downarrow$ & Sliding (m) $\downarrow$ \\
\midrule
With Auxiliary Losses & 2.31 & \textbf{3.47} & \textbf{3.75} & \textbf{0.53} \\
Without Auxiliary Losses & \textbf{2.27} & 5.77 & 6.23 & 0.81 \\
\bottomrule
\end{tabular}%
}
\end{table}

%% file: 6-conclusion.tex
\section{Conclusion}
\label{conclusion}
In this paper, we introduced \name, a novel skeleton-agnostic auto-regressive diffusion model for motion generation that addresses the limitations of existing methods in handling diverse skeletal structures and computational efficiency. By leveraging a U-Net architecture with 1D convolutions for temporal downsampling and upsampling, attention modules for joint and temporal dimensions, cross attention for conditioning on trajectory, and FiLM for conditioning on style and time, our model generates high-quality motion sequences conditioned on style, trajectory, and optional past frames. The use of attention masks based on skeleton topology ensures kinematic consistency, while processing joints in a separate dimension eliminates the need for padding, a common inefficiency in prior work like AnyTop~\cite{anytop}. Our experiments on the \textsc{100style} and \textsc{LAFAN1} datasets demonstrate that \name outperforms state-of-the-art baselines, including CAMDM~\cite{camdm} and MDM~\cite{mdm}. For example, it achieves a lower FID score than MDM, as well as fewer foot penetration frames, reduced sliding distances, and improved trajectory adherence compared to CAMDM. The ability to train on multiple skeleton types simultaneously enables a universal model applicable to diverse datasets, as shown in our superior performance against AnyTop on the combined \textsc{100style} and \textsc{LAFAN1} datasets.

Our ablation studies further validate the importance of key design choices, such as min-max normalization, separate spatial and temporal attention, dataset balancing, and auxiliary losses, which collectively enhance motion quality and training stability. By addressing the computational and structural limitations of transformer-based models, which process full frame sequences, our method offers a scalable and efficient solution for real-world motion synthesis applications. 
At the end, it is worth noting that although \name uses style indices, this can easily be replaced with text input by using a text encoder instead of the style embedding layer.

Looking ahead, future work could explore integrating additional conditioning signals, such as environmental constraints or multi-modal inputs, to further enhance motion realism. Additionally, optimizing our approach for larger datasets with even more diverse skeletons presents promising directions. \name lays a strong foundation for flexible, high-quality motion generation, paving the way for advancements in animation, gaming, and virtual reality.

%% file: 7-figures-only.tex
\input{images/onion_skin}

\input{images/multi_skel}

%% file: images/onion_skin.tex
\begin{figure*}[t]
    \centering
    % First image
    \includegraphics[width=\textwidth]{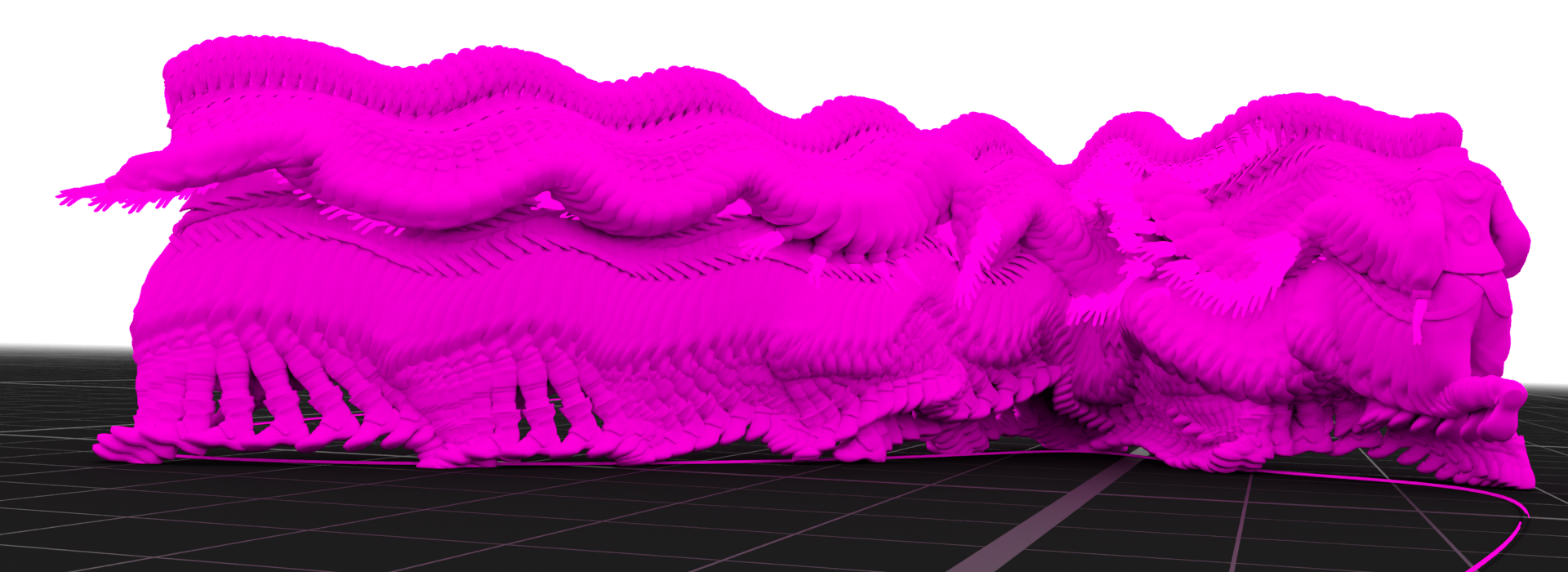}
    % Adding vertical space between images
    \vspace{0.5cm}
    % Second image
    \includegraphics[width=\textwidth]{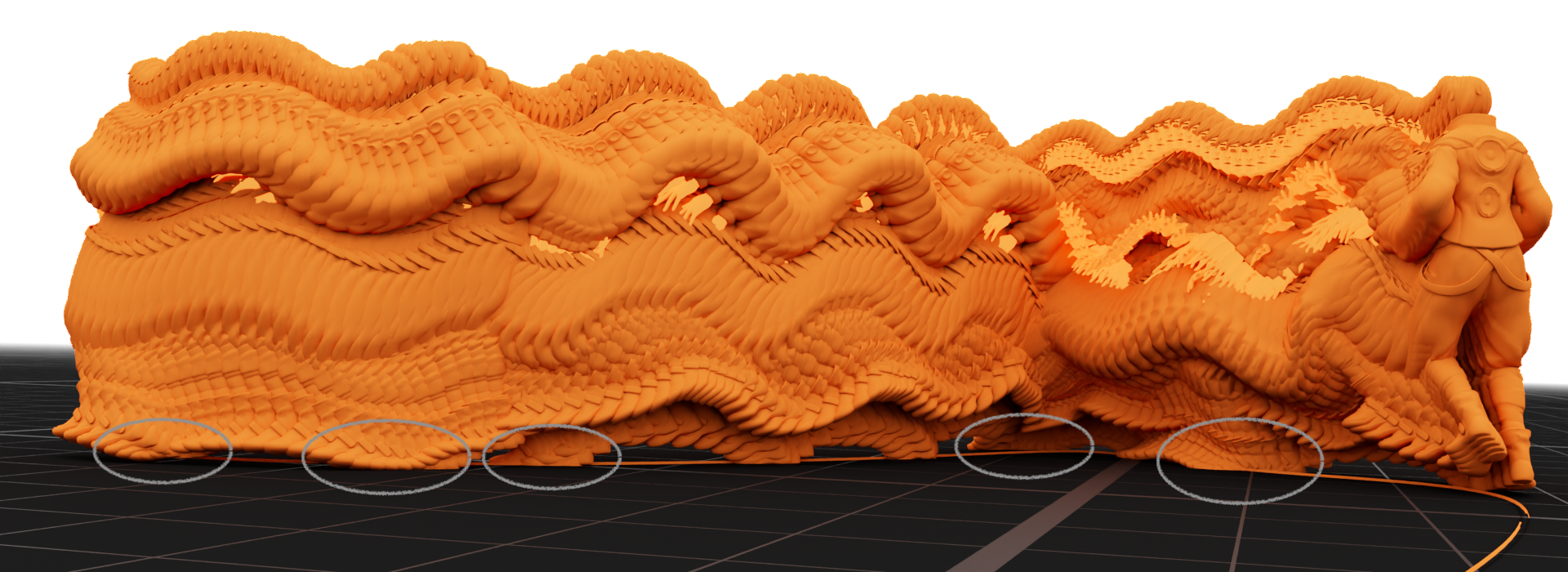}
    % Single caption for the entire figure
    \caption{\textbf{Onion skinning visualization of \name and CAMDM results.} The top and bottom figures compare motion outputs from \name and CAMDM, given the same past frames, style, and trajectory. As shown, our model exhibits noticeably less foot sliding and penetration. These issues are highlighted with ellipses in the CAMDM results for clarity.}
    \label{fig:foot_contact}
\end{figure*}

%% file: images/multi_skel.tex
\begin{figure*}[t]
    \centering
    \includegraphics[width=\linewidth]{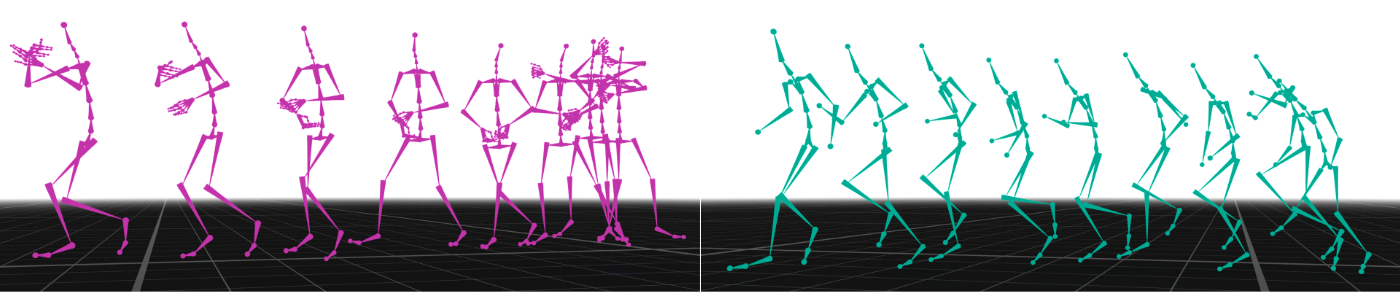}
    \caption{\textbf{Multi-Skeleton Generation.} Left: a motion generated for the skeleton of \textsc{LAFAN1}. Right: a motion generated for the skeleton of \textsc{100Style}. Both the skeletons are generated by the same model, which is trained on the combination the two datasets.}
    \label{fig:multi_skeleton}
    \Description{__}
\end{figure*}